\documentclass[conference]{IEEEtran}
\ifCLASSINFOpdf
  \usepackage[pdftex]{graphicx}
  \DeclareGraphicsExtensions{.pdf}
\else
\fi
\usepackage{amsmath}
\usepackage{amsfonts}
\usepackage{mathtools}
\usepackage[numbers]{natbib}
\usepackage{booktabs}
\usepackage{lipsum}
\usepackage{nicefrac}
\usepackage{soul,color}
\usepackage{algorithm}
\usepackage{algpseudocode}
\newcommand*\diff{\mathop{}\!\mathrm{d}}

\begin{document}
\title{Differentiable Particle Filtering using Optimal Placement Resampling}

\author{\IEEEauthorblockN{Domonkos Csuzdi\IEEEauthorrefmark{1},
Olivér Törő\IEEEauthorrefmark{1} and
Tamás Bécsi\IEEEauthorrefmark{1}}
\IEEEauthorblockA{\IEEEauthorrefmark{1}Department of Control for Transportation and Vehicle Systems, Faculty of Transportation Engineering and Vehicle Engineering,\\
Budapest University of Technology and Economics,
Muegyetem rkp. 3., Budapest, 1111, Hungary\\ Email: domonkos.csuzdi@edu.bme.hu}}

\maketitle

\begin{abstract}
Particle filters are a frequent choice for inference tasks in nonlinear and non-Gaussian state-space models.
They can either be used for state inference by approximating the filtering distribution
or for parameter inference by approximating the marginal data (observation) likelihood.
A good proposal distribution and a good resampling scheme are crucial to obtain low variance estimates.
However, traditional methods like multinomial resampling
introduce nondifferentiability in PF-based loss functions for parameter estimation,
prohibiting gradient-based learning tasks. 
This work proposes a differentiable resampling scheme
by deterministic sampling from an empirical cumulative distribution function.
We evaluate our method on parameter inference tasks and proposal learning.

\end{abstract}

\IEEEpeerreviewmaketitle

\section{Introduction}
Particle filters (PFs) are numeric approximation methods based on Monte Carlo sampling for the general nonlinear Bayesian estimation problem. Monte Carlo methods have been around since the earliest days of computation \cite{mazhdrakov_monte_2018}, with the Metropolis-Hastings algorithm being one of the most notable~\cite{metropolis_equation_1953}. The PF, alternatively referred to as a sequential Monte Carlo estimator, is a recursive method introduced in \cite{gordon_novel_1993}. To evade numerical issues and work efficiently, a PF needs careful design \cite{daum_curse_2003}, meaning the algorithm may include adaptive resampling, regularization, or progressive update. Without at least a resampling step, a PF is ineffective.

Neural networks can replace parts of a PF that are hard to design or parameterize, e.g., the proposal distribution. To train a neural network, one usually applies backpropagation, which is basically a gradient computation. This means that a standard PF structure cannot be used as it almost always involves a nondifferentiable resampling step. Recently, there has been a focus on differentiable particle filters as a consequence.

\subsection{State-Space Models}
\begin{figure}[htb]
    \centering
    \includegraphics[width=0.5\linewidth]{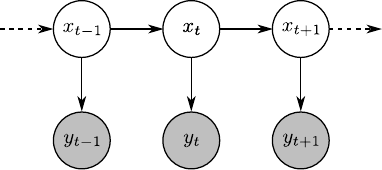}
    \caption{Directed probabilistic graphical model of an SSM. Shaded nodes are observed random variables.}
    \label{fig:pgm-ssm}
\end{figure}

State-space models (SSMs) are a class of sequential models,
where a latent state $x_t$ evolves by a Markov process,
and generates observations $y_t$.
We are interested in SSMs, where there are no external inputs, and the observations are also Markovian, i.e., $y_t$ is conditionally independent of the other states and measurements, given the current state $x_t$.
These relations are shown in Figure \ref{fig:pgm-ssm}.
SSMs can also be represented by a parameterized generative model, defined as
\begin{align}
    x_1 &\sim \pi_\theta(x_1) \, , \\
    x_t &\sim f_\theta(x_t|x_{t-1})\, , \\
    y_t &\sim g_\theta(y_t|x_t)\, ,
\end{align}
where $\pi_\theta(x_1)$ is the initial distribution of the latent state, 
$f_\theta(x_t|x_{t-1})$ is the transition model,
$g_\theta(y_t|x_{t})$ is the observation model, 
and $\theta$ is a parameter set of interest.

There are generally two main goals when working with SSMs: 
performing inference over latent states, 
and performing inference over model parameters $\theta$.
The former has three forms based on which probability density function (pdf) is being computed: $p(x_t|y_{1:t})$ for filtering,
$p(x_{t+1}|y_{1:t})$ for prediction and
$p(x_{t}|y_{1:T})$ for smoothing.
Inference over $\theta$ can be done by maximum likelihood estimation
using the log of the marginal data likelihood:
\begin{align}
\log p_\theta(y_{1:T})&= \log\int p_\theta(x_{1:T},y_{1:T})\mathrm{d}x_{1:T} \\
&=  \sum_{t=1}^{T}\log p_\theta(y_t|y_{1:t-1}) \, ,
\end{align}
where 
\begin{equation}
    p_\theta(y_t|y_{1:t-1}) = \int g_\theta(y_t|x_t)p_\theta(x_t|y_{1:t-1})\mathrm{d}x_t \, .
\end{equation}
Analytical expression for $p_\theta(y_{1:T})$ rarely exists, with a notable exception of 
linear Gaussian SSMs.
Here, the Kalman filter/smoother provides the solution for the state inference tasks,
and also for computing the marginal data log-likelihood.

There are many approximate solutions for the state inference tasks in nonlinear and/or non-Gaussian SSMs,
with the PF being one popular method.
PF can also be used to estimate the marginal data log-likelihood, discussed in the following subsections.

\subsection{Particle Filters}
Particle filters are numeric approximations of the recursive Bayesian estimation problem involving arbitrary density functions \cite{gordon_novel_1993}. A pdf $p$ is represented by $x^i$ particles and their weights $w^i$ in the form of a Dirac sum
\begin{equation} \label{eq:DiracSum}
    \hat{p}(x;\chi, \omega) = \sum\limits_{i=1}^N w^i \delta(x-x^i) \, ,
\end{equation}
where $\chi\coloneqq\{x^i\}_{i=1}^N$ and $\omega\coloneqq\{w^i\}_{i=1}^N$ are the parameters of the approximation $\hat{p}$.

Particles are drawn from the proposal distribution $q$ and then weighted using the measurement likelihood and the transitional density, producing the Dirac sum representation of the posterior distribution. At timestep $t$, the particle approximation of the posterior reads
\begin{equation}
    p(x_t|y_{1:t}) \approx \sum_{i=1}^N w_t^i\delta(x_t-x_t^i) \, ,
\end{equation}
where the posterior weights are
\begin{equation}\label{eq:weights}
    w_t^i =  w_{t-1}^i \frac{p(y_t|x_t^i)p(x_t^i|x_{t-1}^i)}{q(x_t^i|x_{t-1}^i,y_t)} \, .
\end{equation}

In its most basic form, a PF uses the transitional density for the proposal distribution and the likelihood values for the weights. A particle filter to work in practice needs at least an adequate sampling method and a resampling strategy to evade particle depletion \cite{arulampalam_tutorial_2002}
to concentrate particles on regions with high probability mass.
One common resampling scheme is multinomial resampling,
where particles are resampled from a categorical distribution composed
of the normalized particle weights.
For other standard resampling schemes, see \cite{li_resampling_2015}.

Besides posterior inference, PFs can also be used as 
unbiased estimators of the marginal data likelihood if 
the resampling scheme is unbiased (see proofs in \cite{naesseth_variational_2018,maddison_filtering_2017}):
\begin{equation}
    \hat{p}_\theta(y_{1:T}) = \prod_{t=1}^T\sum_{i=1}^N w_t^i,
\end{equation}
\begin{equation}
    \mathbb{E}\left[\hat p_\theta(y_{1:T})\right] = p_\theta(y_{1:T}) \,,
\end{equation}
where the expectation is taken w.r.t. the joint pdf of all the random variables in the particle filtering process.

\subsection{Maximum Likelihood Estimation with Particle Filters}

Maximum likelihood estimation (MLE) in probabilistic models can be formulated as
\begin{equation}
    \hat\theta = \arg \underset{\theta}{\max}\ p_\theta(y_{1:T}) \, .
\end{equation}
Unfortunately for general SSMs, $\hat\theta$ cannot be obtained analytically,
as the marginal data likelihood $p_\theta(y_{1:T})$ is intractable, with the notable exception 
of linear Gaussian SSMs.

A common approach is instead the use of a variational lower bound to the marginal data log-likelihood,
serving as a surrogate objective for optimization.
One typical choice is the evidence lower bound (ELBO) \cite{kingma_auto-encoding_2013}, defined as

\begin{equation}
    \mathcal{L}(y,\theta,\phi) = \mathbb{E}_{q_\phi(x|y)}\left[\log \frac{p_\theta(x,y)}{q_\phi(x|y)}\right] \leq \log p_\theta(y) \, .
\end{equation}
The bound is tight if the variational posterior $q_\phi(x|y)$ is the true posterior $p_\theta(x|y)$.

In time series models, the ELBO can be loose,
as it does not take into consideration the time dependency between states and measurements,
such as defined by an SSM.
The concurrent works of \cite{naesseth_variational_2018,maddison_filtering_2017,le_auto-encoding_2018} propose to use the marginal data log-likelihood estimator of a PF as 
an objective for MLE instead:
\begin{equation}
    \mathcal{L}(y_{1:T},\theta,\phi) = \mathbb{E}\left[\log\prod_{t=1}^T\sum_{i=1}^N w_t^i\right]\label{eq:mco}\approx \frac{1}{B}\sum_{i=1}^B\log\prod_{t=1}^T\sum_{i=1}^N w_t^i,
\end{equation}
where $\theta$ parameterizes the SSM,
and $\phi$ parameterizes the proposal distribution.
Often, $\theta = \phi$.
This bound can potentially be much tighter than ELBO by increasing the number of particles.
The ELBO used in \cite{kingma_auto-encoding_2013} can be extended into
a family of ELBOs, where \eqref{eq:mco} is also an ELBO. 
To avoid notational clutter, we will refer to \eqref{eq:mco} as ELBO 
from now on.

\begin{figure}[htb]
    \centering
    \includegraphics[width=\linewidth]{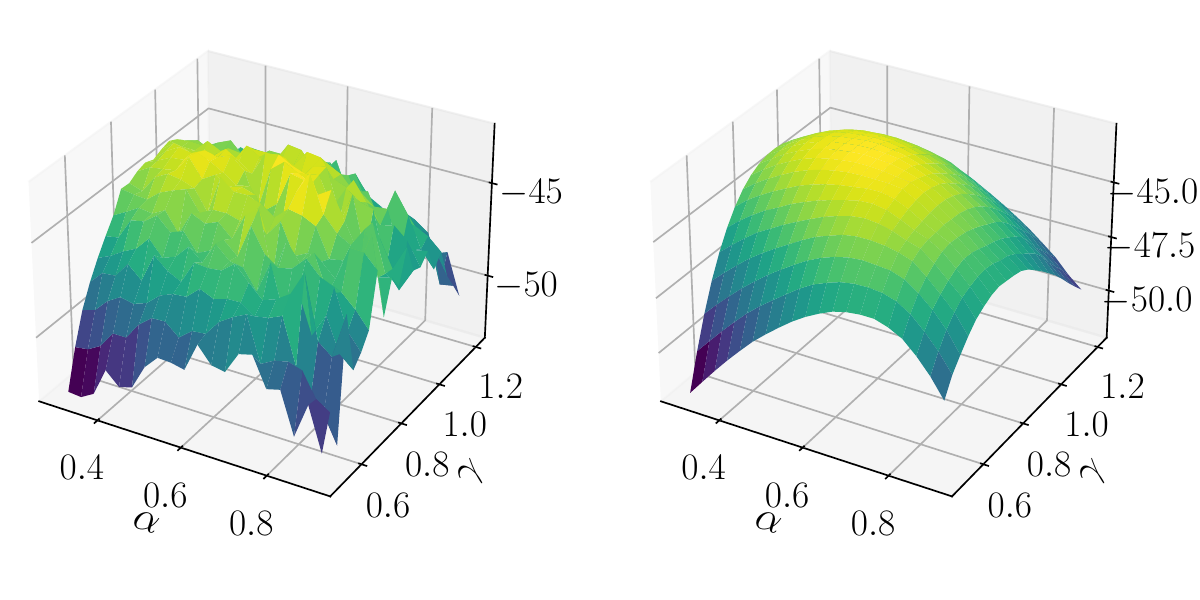}
    \caption{Illustration of the marginal data log-likelihood estimation of a standard PF with multinomial resampling (left), and a differentiable PF with optimal placement resampling (right) for a 1-dimensional linear Gaussian SSM, described in Section \ref{sec:lgssm}.}
    \label{fig:smoothness}
\end{figure}

For optimization via gradient ascent, $\nabla_{\theta,\phi}\mathcal{L}(\cdot)$ is required.
The analytic form of the expectation in \eqref{eq:mco} does not exist; only its Monte Carlo estimator can be obtained.
The expectation is taken w.r.t. the joint distribution of all 
the random variables in the sampling and resampling process, which
may depend on the parameters ($\theta,\phi$).
In most cases, the reparameterization trick \cite{kingma_auto-encoding_2013} can be
used to move the gradient computation inside the expectation.
This requires the ability to reparameterize the proposal distribution,
and the resampling distribution.
The problems arise with the latter, as traditional resampling schemes,
like multinomial resampling, are discontinuous w.r.t. the model parameters.
This means that small changes in the model parameters can lead to abrupt changes in the resampling outcome and thus in the Monte Carlo estimate \eqref{eq:mco}, illustrated in Figure \ref{fig:smoothness}. For instance, a particle with a low weight may not be resampled initially, but slight adjustments in the model parameters could cause its weight to increase, resulting in its occasional resampling.
Ignoring this effect leads to a high-variance Monte Carlo estimate of the gradient,
and prevents backpropagation through time.

\section{Related Work and Contributions}
Performing joint model and proposal learning
in directed probabilistic models using stochastic gradient ascent
goes back to \cite{kingma_auto-encoding_2013}.
Their approach was further extended by
importance sampling principles in \cite{burda_importance_2015}.
Although these methods also work on structured time series models,
the authors in \cite{naesseth_variational_2018,maddison_filtering_2017,le_auto-encoding_2018}
concurrently showed that the marginal likelihood estimator
of a PF provides a tighter lower bound for such models.
These works also noted the fact that the resampling step
is nondifferentiable but did not propose a solution,
instead, they used a biased gradient estimator.
The nondifferentiable problem also came up in
\cite{karkus_particle_2018,jonschkowski_differentiable_2018},
where the aim was to learn motion and measurement models
for robotic localization, parameterized by neural networks and using a
PF-based supervised loss. 
Since then, multiple differentiable PF techniques have been proposed
\cite{karkus_particle_2018,scibior_differentiable_2021,younis_differentiable_2023,singh_particle-based_2023, zhu_towards_2020, rosato_efficient_2022},
but most notably \cite{corenflos_differentiable_2021},
which was the first fully differentiable particle filter.
A recent overview of differentiable particle filters is presented in \cite{chen_overview_2023},
while \cite{kloss_how_2021} provides guidelines on training them.

In this paper, we propose a differentiable particle filter structure. Our work is similar to \cite{malik_particle_2011}, where they also make the
likelihood estimation of a PF continuous by sampling from the cdf.
Contrary to their work, we use deterministic sampling,
which satisfies a given optimality criteria. 
We provide empirical results in learning
system and proposal parameters on 
toy examples and also on real-world data.
\section{Deterministic Sampling}
\begin{figure*}[t]
        \centering\includegraphics[width=0.8\linewidth]{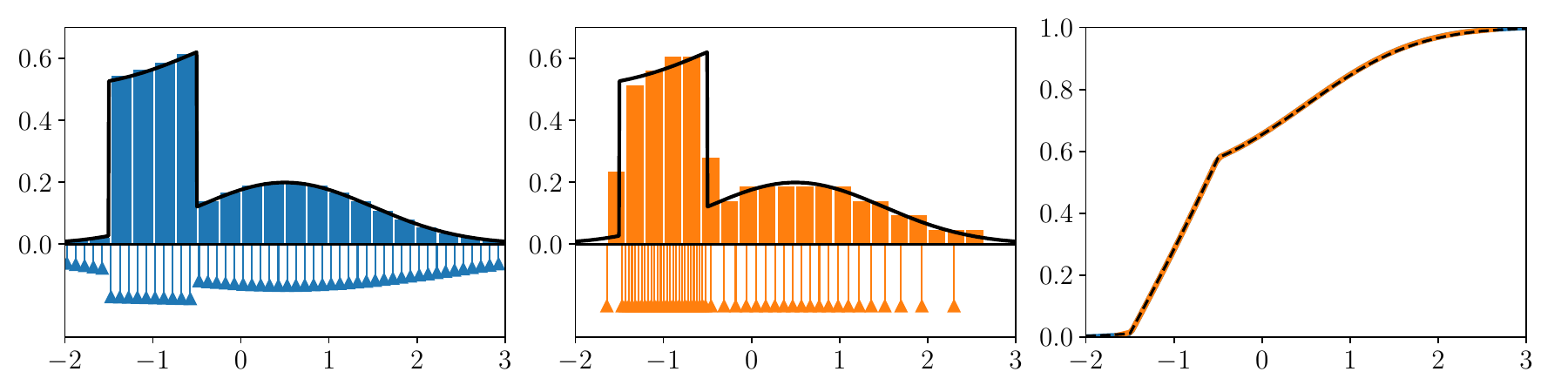}
        \centering\includegraphics[width=0.8\linewidth]{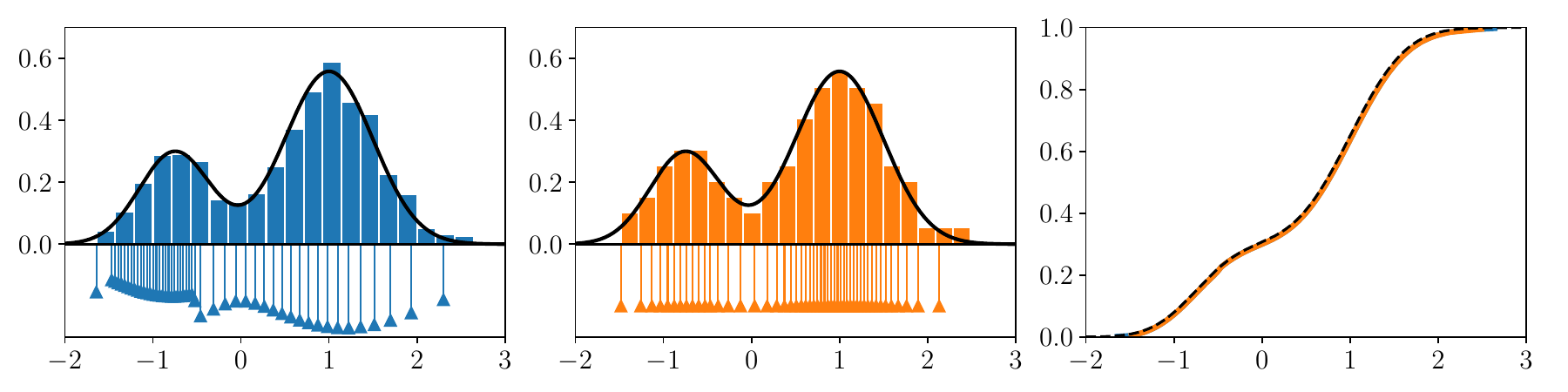}
        \caption{Illustration of optimal placement resampling.
        The goal is to perform resampling, i.e., create an unweighted particle set from a weighted particle set~\eqref{eq:DiracSum} by moving particles to positions with high probability mass.
        The true pdf (black) and true cdf (dashed black) do not change during resampling, only their representation by particles.
        The two rows show two consecutive timesteps.
        Left: weighted particle set before resampling, with stem length proportional to particle weight. Middle: resampled particle set by optimal placement resampling.
        Right: empirical cdfs.
        In the second row, the resampled particles in the previous timestep are weighted according to \eqref{eq:weights}, and the resampling is performed again.
        Due to the particles' finite representational power, their histograms slightly differ before and after resampling.
        \label{fig:nagyFig}
}
\end{figure*}%
\begin{figure}
    \centering
    \includegraphics[width=1\linewidth]{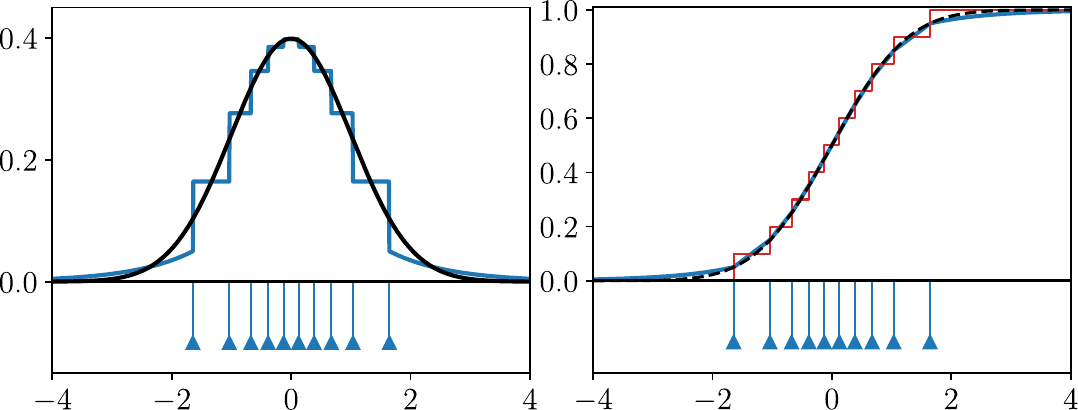}
    \caption{Illustration of empirical pdf (left) and cdf (right) based on the optimal particle locations. Black line: analytical pdf and cdf. Blue line: empirical pdf and cdf. Blue triangles: particle locations. Red line: traditional step-wise empirical cdf, which we replaced by the blue line approximation given in \eqref{eq:empCDF}.}
    \label{fig:empiric}
\end{figure}
Resampling in PFs prevents particle degeneration by eliminating low-weight particles and moving particles to highly-weighted particle positions. 
This procedure essentially replaces the weighted Dirac mixture representation of the posterior with an unweighted Dirac mixture representation.
It is important to mention that the underlying posterior pdf does not change,
only its representation by the particles.
Standard resampling schemes have two main properties: they are stochastic, and they do not introduce new particle positions.

Schrempf et al. \cite{schrempf_density_2006} proposed a deterministic alternative
for approximating a pdf by unweighted particles. Their approach builds on the integral quadratic distance
\begin{equation} \label{eq:DistMeasure}
    D = \int_{-\infty}^{\infty} \left( F(x) - G(x) \right)^2 \diff{x} \, ,
\end{equation}
where $F$ and $G$ are the cumulative density functions of the distributions involved.
If a distribution is approximated by particles in the form \eqref{eq:DiracSum}, the corresponding empirical cdf is
\begin{align} \label{eq:Fhat}
    \hat{F}(x;\chi,\omega) &= \int_{-\infty}^x \hat{f}(s;\chi, \omega) \diff{s}=\sum_{i=1}^N w^i H(x-x^i) \, ,
\end{align}
where $H$ is the Heaviside function defined as
\begin{equation}
    H(x) = \left\{ \begin{array}{cl}
         0 \, , & x<0 \\
         \nicefrac{1}{2} \, , & x=0 \\
         1 \, , & x>0
    \end{array} \right. \, .
\end{equation}

The approach proposed in \cite{schrempf_density_2006} is the following. Let the particle approximation use equal weights, i.e., $w^i=1/N$. Inserting an exact cdf $F$ and its particle approximation $\hat{F}$ from \eqref{eq:Fhat} into the distance function \eqref{eq:DistMeasure} and minimizing it in terms of the $x^i$ particle positions one finds that
\begin{equation} \label{eq:Optimal}
    F(x^i) = \frac{2i-1}{2N} \, \,  (i=1\dots N) \, ,
\end{equation}
which is an implicit relation that gives the optimal $x^i$ positions of the particles to represent the exact distribution $F$ while minimizing the integral squared distance \eqref{eq:DistMeasure} between $F$ and $\hat{F}$.

To compute the $x^i$ positions, the cdf must be inverted, which is not generally feasible. For the uniform distribution on the interval [0,1], the solution is
\begin{equation}
    x^i = \frac{2i-1}{2N} \, .
\end{equation}
% while for the standard normal distribution, using the  inverse error function $\mathrm{erf}^{-1}$, it is
% \begin{equation}
%     x^i = \sqrt{2}\mathrm{erf}^{-1}\left(\frac{2i-1-N}{N}\right) \, .
% \end{equation}

For an arbitrary pdf $f(x)$, Schrempf et al. introduced a homotopy progression from an initial pdf $g(x)$ to $f(x)$, which generates a flow of the particles to occupy the optimal positions.

\subsection{Optimal Placement Resampling}
Instead of using a flow, we propose to place the particles directly to the optimal positions according to \eqref{eq:Optimal}. To this end, we create a cdf approximation constructed from particle weights and positions using easily invertible functions.

Consider a weighted particle representation of a distribution with particle positions $x^i$ and weights $w^i$. The pdf will be approximated, instead of a Dirac sum, with a weighted Heaviside sum in the form
\begin{equation} \label{eq:empPDF}
    f(x) \!=\! \left\{\!\! \begin{array}{ll}
    \!\dfrac{w^1}{2}\exp(x-x^1) \, , & x \leq x^1 \\[0.5em]
         \!\varpi^1 + \sum_{i=2}^{N-1}  (\varpi^i\!-\!\varpi^{i-1})H(x\!-\!x^i) , \!\!\! &  x^1 \!<\! x \!<\! x^N \\[0.2em]
         \!\dfrac{w^N}{2}\exp(-x+x^N) ,& x^N\leq x
    \end{array}  \right.
\end{equation}
where
\begin{equation}
    \varpi^i = \frac{1}{2}\frac{w^i + w^{i-1}}{x^i - x^{i-1}} \, ,
\end{equation}
which is also the steepness of the ramp function in the cdf.

The ramp function is defined as
\begin{equation}
    R(x) = \int_{-\infty}^x H(s) \diff{s} \, ,
\end{equation}
which allows writing
\begin{equation} \label{eq:empCDF}
    F(x) = \left\{ \begin{array}{ll}
            \dfrac{w^1}{2}\exp(x-x^1) \, , &x \leq x^1 \\[0.9em]
         \begin{aligned} \varpi^1&R(x-x^1) +\\
         &\sum_{i=2}^{N-1} (\varpi^i\!-\!\varpi^{i-1}) R(x-x^i) \, , \end{aligned} &x^1 < x < x^N \\[0.9em]
         1-\frac{w^N}{2}\exp(-x+x^N) \, , &x^N\leq x
    \end{array}  \right.
\end{equation}

The exponential leading and trailing parts in the piecewise definitions \eqref{eq:empPDF} and \eqref{eq:empCDF} ensure that these functions can be evaluated at any point and insert a minimal amount of arbitrary probability mass to regions that are not covered by particles (Fig. \ref{fig:empiric}).

To be able to use the empirical cdf for computing the optimal particle positions in \eqref{eq:Optimal}, the inverse cdf is needed, which consists of linear and logarithmic terms:
\begin{equation} \label{eq:invEmpCDF}
    F^{-1}(w) = \left\{ \begin{array}{ll}
        x^1 + \log(\frac{2w}{w^1}) \, , & w\leq w^1 \\[0.5em]
        \sum_{i=2}^{N-1} R(w-w^i)/\varpi^i \, , &  w^1 < w < x^N \\[0.5em]
        x^N + \log(\frac{w^N}{2-2w}) \, , & w^N \leq w
    \end{array} \right.
\end{equation}

We will use the ideas presented above. In the PF update step, no stochastic resampling is performed after the weighted particle set is obtained. Instead, each particle will move deterministically to the optimal position according to \eqref{eq:Optimal} using the inverse cdf constructed as in \eqref{eq:invEmpCDF}. Note that the particles have to be sorted in an increasing order. We will call this scheme optimal placement resampling (OPR).
OPR obtains optimally distributed samples from the weighted empirical posterior distribution,
thus not having duplicate particles, maintaining diversity in the particle set. However, high-probability regions in the posterior still concentrate more particles. Figure \ref{fig:nagyFig}. illustrates the process.

\section{Empirical Study}

\subsection{Linear Gaussian State-Space Model}\label{sec:lgssm}
For our first study, we examine a one-dimensional linear Gaussian state-space model (LGSSM), defined as
\begin{align}
    x_t &= \alpha x_{t-1} + v_t\label{eq:lgssm-x} \\
    y_t &= \gamma x_t + e_t\label{eq:lgssm-y} \, , 
\end{align}
with $v_t\sim\mathcal{N}(0,\sigma_x^2)$,
$e_t\sim\mathcal{N}(0,\sigma_y^2)$ and
$x_1\sim\mathcal{N}(0,\sigma_x^2)$.
An analytic expression of $p_\theta(y_{1:T})$
exists for this simple model, and the true
marginal data log-likelihood can be computed using the Kalman filter.
We generated a  measurement dataset of $T=100$ steps,
using $\theta^* = [\alpha^*, \gamma^*] = [0.5,1], \sigma_x^2 = 0.3,
\sigma_y^2 = 0.1$. 
We then tried to learn the maximum likelihood estimation of the parameters $\theta = [\alpha,\gamma]$ using stochastic gradient ascent on the approximation of the ELBO from
a PF with multinomial resampling (PF-MR) and optimal placement resampling (PF-OPR).
The particle filters were run with $N=50$ particles and in $B=50$ batches to evaluate
the ELBO \eqref{eq:mco}.
The Adam optimizer  \cite{kingma_adam_2017} was used with
a learning rate of 0.01 for 200 epochs,
and $\theta_0 = [1,1.5]$ as initial parameters.
Similarly to the result in  \cite{corenflos_differentiable_2021},
we obtain that, in this simplistic case, the nondifferentiability
of PF-MR does not cause problems in learning, and the results compared to PF-OPR are similar. The ELBO from both PF methods has a relative error of 1.5$\%$ to the true marginal data log-likelihood.

\subsection{Proposal Distribution Learning}

In particle filters, the proposal distribution is a key choice.
Based on the experiments in \cite{naesseth_variational_2018},
we learn a time-varying proposal distribution in the form of
\begin{equation}
    r_\lambda(x_t|x_{t-1}) = \mathcal{N}(\mu_t + \beta_t\alpha x_{t-1},\sigma_t^2)
\end{equation}
for
the LGSSM defined in \eqref{eq:lgssm-x}-\eqref{eq:lgssm-y},
with $\lambda = \{\mu_t,\beta_t, \sigma^2_t\}_{t=1}^T$.
As the parameters are time-varying in this case, efficient learning requires the ability to backpropagate through time. 

\begin{figure}[t]
    \centering
    \includegraphics[width=\linewidth]{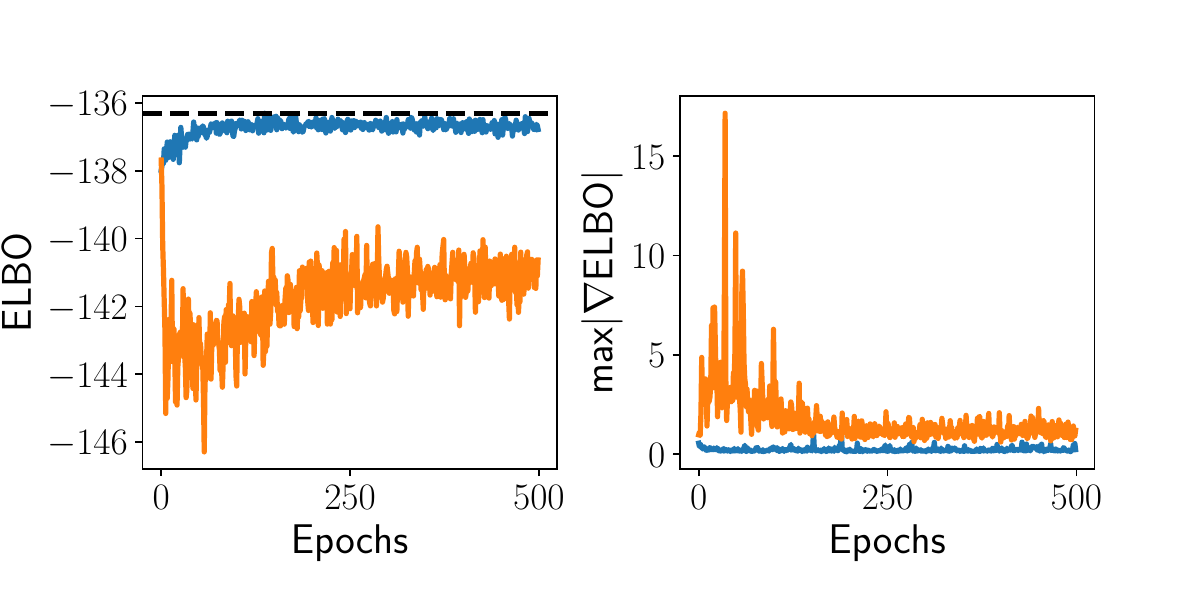}
    \caption{Proposal distribution learning for an LGSSM using PF-MR (orange)
    and PF-OPR (blue). Higher ELBO is better. }
    \label{fig:prop-learning}
\end{figure}

Using $\alpha = 0.42, \gamma = 1, \sigma^2_x = 1, \sigma^2_y = 0.1$,
we performed the maximum likelihood learning of the proposal parameters
using the ELBO estimate from PF-MR and PF-OPR.
The Adam optimizer was chosen with a learning rate of 0.1,
with initial parameters $\mu_t = 0, \beta_t=1, \log\sigma_t = 0$ for $t=1...T$.
The particle filters were run in $B=50$ batches with $N=100$ particles.
Figure \ref{fig:prop-learning} shows the ELBO estimate of PF-MR in orange
and PF-OPR in blue, along with the maximal component of the 
gradient of the ELBO
w.r.t $\lambda$.
It is clearly apparent that multinomial resampling
performs worse for this task, which is likely due to
its inability to perform backpropagation through time 
as its resampling scheme is nondifferentiable.
The execution time per epoch was 83.4 ms for PF-MR
and 113.7 ms for PF-OPR.
We note that the most time-consuming process in PF-OPR
is sorting particles required for 
constructing the cdf. However, the OPR algorithm still runs
in $\mathcal{O}(N)$ time complexity.

\subsection{Stochastic Volatility Model}

\begin{figure}[t]
    \centering
    \includegraphics[width=\linewidth]{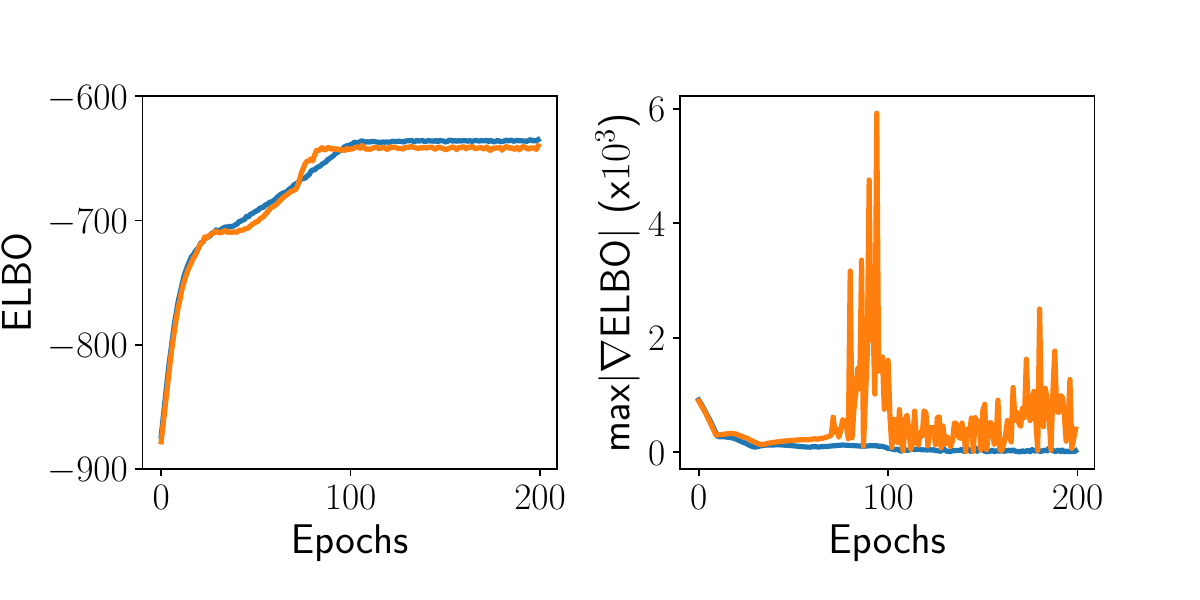}
    \caption{The estimated ELBO and its maximum gradient component w.r.t. $\theta$ for a parameter inference task in a stochastic volatility model. Blue: PF-OPR, orange: PF-MR. Higher ELBO is better.}
    \label{fig:stochvol}
\end{figure}

The stochastic volatility model \cite{hull_pricing_1987} is a common choice for modeling
financial processes, like price changes in the stock market.  
The model is defined as 
\begin{align}
    x_t &= \mu + \phi(x_{t-1} - \mu) + v_t \\
    y_t &= \exp\left(\frac{x_t}{2}\right)e_t \, ,
\end{align}
with $v_t \sim \mathcal{N}(0,\sigma_x^2)$, $e_t \sim \mathcal{N}(0,\sigma_y^2)$, 
$x_1 \sim \mathcal{N}\left(\mu,\frac{\sigma_x^2}{1-\phi^2}\right)$.
The parameter set of interest is $\theta = [\mu,\phi,\sigma_x,\sigma_y]$.
The observations are the log-returns, which are defined as
\begin{equation}
    y_t = 100\log\left(\frac{s_t}{s_{t-1}}\right) \, ,
\end{equation}
where $s_t$ is the price at $t$.
Our observation dataset is constructed from daily
EUR/HUF exchange rates of the European Central Bank from
01.01.2017 to 01.01.2023. 
As the analytic expression for $p_\theta(y_{1:T})$ is not 
tractable, we used the approximates produced by PF-MR and PF-OPR.
The training results with $N=50$ particles, $B=50$ batches, and using the Adam optimizer with a learning rate of $0.01$
is shown in Figure \ref{fig:stochvol}.
The ELBO for PF-OPR is -634.9, and the ELBO for PF-MR is -640.0,
meaning that PF-OPR is capable of providing a tighter and 
therefore better ELBO estimate.

\section{Conclusions and Future Work}
We introduced optimal placement resampling for particle filters
as a way of solving the nondifferentiability problem of standard resampling
algorithms. 
Our method uses optimally placed samples from a hand-crafted empirical
cumulative distribution function based on the prior weighted particle set.
We empirically showed that optimal placement resampling outperforms
the nondifferentiable multinomial resampling in model and proposal learning
tasks.

In the current form, the optimal placement resampling works only in one dimension due to the fact that it relies on the cdf. The cdf is not a uniquely defined function since the relation in $F(x) = P(X\leq x)$ is arbitrary. It does not cause a problem in one dimension because the cdfs $P(X\leq x)$ and $P(X\geq x)$ lead to the same results. Starting from dimension two, this invariance no longer holds, and alternatives to the conventional cdf have to be used \cite{hanebeck_localized_2008}. A future task is to design easily computable optimal placement strategies in multiple dimensions using either an alternative cdf or another placement strategy.

\section*{Acknowledgment}
This research was supported by the European Union within the framework of the National Laboratory for Autonomous Systems. (RRF-2.3.1-21-2022-00002).

DC was supported by ÚNKP-23-3-I-BME-146 New National Excellence Program of the Ministry for Culture and Innovation.

The authors would also like to thank Adrien Corenflos and James Thornton for making their code in \cite{corenflos_differentiable_2021} publicly available.
\bibliographystyle{IEEEtran}

\begin{thebibliography}{10}
\providecommand{\url}[1]{#1}
\csname url@samestyle\endcsname
\providecommand{\newblock}{\relax}
\providecommand{\bibinfo}[2]{#2}
\providecommand{\BIBentrySTDinterwordspacing}{\spaceskip=0pt\relax}
\providecommand{\BIBentryALTinterwordstretchfactor}{4}
\providecommand{\BIBentryALTinterwordspacing}{\spaceskip=\fontdimen2\font plus
\BIBentryALTinterwordstretchfactor\fontdimen3\font minus
  \fontdimen4\font\relax}
\providecommand{\BIBforeignlanguage}[2]{{%
\expandafter\ifx\csname l@#1\endcsname\relax
\typeout{** WARNING: IEEEtran.bst: No hyphenation pattern has been}%
\typeout{** loaded for the language `#1'. Using the pattern for}%
\typeout{** the default language instead.}%
\else
\language=\csname l@#1\endcsname
\fi
#2}}
\providecommand{\BIBdecl}{\relax}
\BIBdecl

\bibitem{mazhdrakov_monte_2018}
P.~M. Mazhdrakov, D.~Benov, and P.~N. Valkanov,
  \emph{\BIBforeignlanguage{English}{The {Monte} {Carlo} {Method}}}.\hskip 1em
  plus 0.5em minus 0.4em\relax ACMO Academic Press, Aug. 2018.

\bibitem{metropolis_equation_1953}
N.~Metropolis, A.~W. Rosenbluth, M.~N. Rosenbluth, A.~H. Teller, and E.~Teller,
  ``\BIBforeignlanguage{en}{Equation of {State} {Calculations} by {Fast}
  {Computing} {Machines}},'' \emph{\BIBforeignlanguage{en}{The Journal of
  Chemical Physics}}, vol.~21, no.~6, pp. 1087--1092, Jun. 1953.

\bibitem{gordon_novel_1993}
N.~Gordon, D.~Salmond, and A.~Smith, ``\BIBforeignlanguage{en}{Novel approach
  to nonlinear/non-{Gaussian} {Bayesian} state estimation},''
  \emph{\BIBforeignlanguage{en}{IEE Proceedings F Radar and Signal
  Processing}}, vol. 140, no.~2, p. 107, 1993.

\bibitem{daum_curse_2003}
F.~Daum and J.~Huang, ``Curse of dimensionality and particle filters,'' in
  \emph{2003 {IEEE} {Aerospace} {Conference} {Proceedings} ({Cat}.
  {No}.{03TH8652})}, vol.~4, Mar. 2003, pp. 4\_1979--4\_1993, iSSN: 1095-323X.

\bibitem{arulampalam_tutorial_2002}
M.~Arulampalam, S.~Maskell, N.~Gordon, and T.~Clapp, ``A tutorial on particle
  filters for online nonlinear/non-{Gaussian} {Bayesian} tracking,'' \emph{IEEE
  Transactions on Signal Processing}, vol.~50, no.~2, pp. 174--188, Feb. 2002.

\bibitem{li_resampling_2015}
T.~Li, M.~Bolic, and P.~M. Djuric, ``\BIBforeignlanguage{en}{Resampling
  {Methods} for {Particle} {Filtering}: {Classification}, implementation, and
  strategies},'' \emph{\BIBforeignlanguage{en}{IEEE Signal Processing
  Magazine}}, vol.~32, no.~3, pp. 70--86, May 2015.

\bibitem{naesseth_variational_2018}
C.~Naesseth, S.~Linderman, R.~Ranganath, and D.~Blei,
  ``\BIBforeignlanguage{en}{Variational {Sequential} {Monte} {Carlo}},'' in
  \emph{\BIBforeignlanguage{en}{Proceedings of the {Twenty}-{First}
  {International} {Conference} on {Artificial} {Intelligence} and
  {Statistics}}}.\hskip 1em plus 0.5em minus 0.4em\relax PMLR, Mar. 2018, pp.
  968--977, iSSN: 2640-3498.

\bibitem{maddison_filtering_2017}
C.~J. Maddison, J.~Lawson, G.~Tucker, N.~Heess, M.~Norouzi, A.~Mnih, A.~Doucet,
  and Y.~Teh, ``Filtering {Variational} {Objectives},'' in \emph{Advances in
  {Neural} {Information} {Processing} {Systems}}, vol.~30.\hskip 1em plus 0.5em
  minus 0.4em\relax Curran Associates, Inc., 2017.

\bibitem{kingma_auto-encoding_2013}
D.~P. Kingma and M.~Welling, ``Auto-{Encoding} {Variational} {Bayes},''
  \emph{CoRR}, Dec. 2013.

\bibitem{le_auto-encoding_2018}
T.~Le, M.~Igl, T.~Rainforth, T.~Jin, and F.~Wood,
  ``\BIBforeignlanguage{en}{Auto-encoding sequential {Monte} {Carlo}},''
  \emph{\BIBforeignlanguage{en}{International Conference on Learning
  Representations (ICLR)}}, 2018.

\bibitem{burda_importance_2015}
Y.~Burda, R.~Grosse, and R.~Salakhutdinov, ``\BIBforeignlanguage{en}{Importance
  {Weighted} {Autoencoders}},'' \emph{\BIBforeignlanguage{en}{International
  Conference on Learning Representations}}, 2015.

\bibitem{karkus_particle_2018}
P.~Karkus, D.~Hsu, and W.~S. Lee, ``\BIBforeignlanguage{en}{Particle {Filter}
  {Networks} with {Application} to {Visual} {Localization}},'' in
  \emph{\BIBforeignlanguage{en}{Proceedings of {The} 2nd {Conference} on
  {Robot} {Learning}}}.\hskip 1em plus 0.5em minus 0.4em\relax PMLR, Oct. 2018,
  pp. 169--178, iSSN: 2640-3498.

\bibitem{jonschkowski_differentiable_2018}
R.~Jonschkowski, D.~Rastogi, and O.~Brock,
  ``\BIBforeignlanguage{en}{Differentiable {Particle} {Filters}: {End}-to-{End}
  {Learning} with {Algorithmic} {Priors}},'' in
  \emph{\BIBforeignlanguage{en}{Robotics: {Science} and {Systems}
  {XIV}}}.\hskip 1em plus 0.5em minus 0.4em\relax Robotics: Science and Systems
  Foundation, Jun. 2018.

\bibitem{scibior_differentiable_2021}
A.~Ścibior and F.~Wood, ``Differentiable {Particle} {Filtering} without
  {Modifying} the {Forward} {Pass},'' Oct. 2021, arXiv:2106.10314 [cs, stat].

\bibitem{younis_differentiable_2023}
A.~Younis and E.~Sudderth, ``\BIBforeignlanguage{en}{Differentiable and
  {Stable} {Long}-{Range} {Tracking} of {Multiple} {Posterior} {Modes}},''
  \emph{\BIBforeignlanguage{en}{Advances in Neural Information Processing
  Systems}}, vol.~36, Dec. 2023.

\bibitem{singh_particle-based_2023}
A.~Singh, O.~Makhlouf, M.~Igl, J.~Messias, A.~Doucet, and S.~Whiteson,
  ``\BIBforeignlanguage{en}{Particle-{Based} {Score} {Estimation} for {State}
  {Space} {Model} {Learning} in {Autonomous} {Driving}},'' in
  \emph{\BIBforeignlanguage{en}{Proceedings of {The} 6th {Conference} on
  {Robot} {Learning}}}.\hskip 1em plus 0.5em minus 0.4em\relax PMLR, Mar. 2023,
  pp. 1168--1177, iSSN: 2640-3498.

\bibitem{zhu_towards_2020}
M.~Zhu, K.~Murphy, and R.~Jonschkowski, ``Towards {Differentiable}
  {Resampling},'' Apr. 2020, arXiv:2004.11938 [cs, stat].

\bibitem{rosato_efficient_2022}
C.~Rosato, L.~Devlin, V.~Beraud, P.~Horridge, T.~B. Schön, and S.~Maskell,
  ``Efficient {Learning} of the {Parameters} of {Non}-{Linear} {Models} {Using}
  {Differentiable} {Resampling} in {Particle} {Filters},'' \emph{IEEE
  Transactions on Signal Processing}, vol.~70, pp. 3676--3692, 2022.

\bibitem{corenflos_differentiable_2021}
A.~Corenflos, J.~Thornton, G.~Deligiannidis, and A.~Doucet,
  ``\BIBforeignlanguage{en}{Differentiable {Particle} {Filtering} via
  {Entropy}-{Regularized} {Optimal} {Transport}},'' in
  \emph{\BIBforeignlanguage{en}{Proceedings of the 38th {International}
  {Conference} on {Machine} {Learning}}}.\hskip 1em plus 0.5em minus
  0.4em\relax PMLR, Jul. 2021, pp. 2100--2111, iSSN: 2640-3498.

\bibitem{chen_overview_2023}
X.~Chen and Y.~Li, ``\BIBforeignlanguage{en}{An overview of differentiable
  particle filters for data-adaptive sequential {Bayesian} inference},''
  \emph{\BIBforeignlanguage{en}{Foundations of Data Science}}, Dec. 2023.

\bibitem{kloss_how_2021}
A.~Kloss, G.~Martius, and J.~Bohg, ``\BIBforeignlanguage{en}{How to train your
  differentiable filter},'' \emph{\BIBforeignlanguage{en}{Autonomous Robots}},
  vol.~45, no.~4, pp. 561--578, May 2021.

\bibitem{malik_particle_2011}
S.~Malik and M.~K. Pitt, ``Particle filters for continuous likelihood
  evaluation and maximisation,'' \emph{Journal of Econometrics}, vol. 165,
  no.~2, pp. 190--209, Dec. 2011.

\bibitem{schrempf_density_2006}
O.~C. Schrempf, D.~Brunn, and U.~D. Hanebeck, ``\BIBforeignlanguage{en}{Density
  {Approximation} {Based} on {Dirac} {Mixtures} with {Regard} to {Nonlinear}
  {Estimation} and {Filtering}},'' in \emph{\BIBforeignlanguage{en}{Proceedings
  of the 45th {IEEE} {Conference} on {Decision} and {Control}}}.\hskip 1em plus
  0.5em minus 0.4em\relax San Diego, CA, USA: IEEE, 2006, pp. 1709--1714.

\bibitem{kingma_adam_2017}
D.~P. Kingma and J.~Ba, ``Adam: {A} {Method} for {Stochastic} {Optimization},''
  Jan. 2017, arXiv:1412.6980 [cs].

\bibitem{hull_pricing_1987}
J.~Hull and A.~White, ``\BIBforeignlanguage{en}{The {Pricing} of {Options} on
  {Assets} with {Stochastic} {Volatilities}},''
  \emph{\BIBforeignlanguage{en}{The Journal of Finance}}, vol.~42, no.~2, pp.
  281--300, 1987.

\bibitem{hanebeck_localized_2008}
U.~D. Hanebeck and V.~Klumpp, ``Localized {Cumulative} {Distributions} and a
  multivariate generalization of the {Cramér}-von {Mises} distance,'' in
  \emph{2008 {IEEE} {International} {Conference} on {Multisensor} {Fusion} and
  {Integration} for {Intelligent} {Systems}}, Aug. 2008, pp. 33--39.

\end{thebibliography}

\end{document}